\documentclass[conference]{IEEEtran}

\usepackage[utf8]{inputenc}
\usepackage[T1]{fontenc}
\usepackage{cite}
\usepackage{amsmath,amssymb,amsfonts}
\usepackage{graphicx}
\usepackage{booktabs}
\usepackage{url}
\usepackage{hyperref}
\usepackage{listings}
\usepackage{xcolor}
\usepackage{multirow}

\lstdefinestyle{json}{
    basicstyle=\ttfamily\footnotesize,
    breaklines=true,
    frame=single,
    columns=fullflexible,
    showstringspaces=false
}

\title{Environment-Grounded Multi-Agent Workflow for Autonomous Penetration Testing}
\author{

\IEEEauthorblockN{Michael Somma\IEEEauthorrefmark{1}\IEEEauthorrefmark{3},
Markus Großpointner\IEEEauthorrefmark{1}\IEEEauthorrefmark{2},
Paul Zabalegui\IEEEauthorrefmark{4},
Eppu Heilimo\IEEEauthorrefmark{5},
Branka Stojanović\IEEEauthorrefmark{1}}

\IEEEauthorblockA{
\IEEEauthorrefmark{1}JOANNEUM RESEARCH Forschungsgesellschaft mbH, DIGITAL – Institute for Digital Technologies\\
Steyrergasse 17, 8010 Graz, Austria\\
Email: {michael.somma@joanneum.at, markus.grosspointner@joanneum.at, branka.stojanovic@joanneum.at}
}

\IEEEauthorblockA{
\IEEEauthorrefmark{3}TU Graz, Institute for Technical Informatics\\
Inffeldgasse 16/I, 8010 Graz, Austria
}

\IEEEauthorblockA{
\IEEEauthorrefmark{4}Alias Robotics S.L.\\
Calle Venta de la Estrella, 6 Pab. 130\\
01006 Vitoria-Gasteiz, Álava, Spain\\
Email: {paul@aliasrobotics.com}
}

\IEEEauthorblockA{
\IEEEauthorrefmark{5}Jamk University of Applied Sciences, Institute of Information Technology\\
40101 Jyväskylä, Finland\\
Email: {eppu.heilimo@jamk.fi}
}
}

\begin{document}
\maketitle

% ---------- Abstract ---------- TBD
\begin{abstract}
The increasing complexity and interconnectivity of digital infrastructures make scalable and reliable security assessment methods essential. Robotic systems represent a particularly important class of operational technology, as modern robots are highly networked cyber-physical systems deployed in domains such as industrial automation, logistics, and autonomous services. This paper explores the use of large language models for automated penetration testing in robotic environments. We propose an environment-grounded multi-agent architecture tailored to Robotics-based systems. The approach dynamically constructs a shared graph-based memory during execution that captures the observable system state, including network topology, communication channels, vulnerabilities, and attempted exploits. This enables structured automation while maintaining traceability and effective context management throughout the testing process. Evaluated across multiple iterations within a specialized robotics Capture-the-Flag scenario (ROS/ROS2), the system demonstrated high reliability, successfully completing the challenge in 100\% of test runs (n=5). This performance significantly exceeds literature benchmarks while maintaining the traceability and human oversight required by frameworks like the EU AI Act.
\end{abstract}


% ==================================================
\section{Introduction} 

As digital infrastructures become increasingly complex and interconnected, 
the need for reliable, scalable, and efficient security assessment methods 
is growing rapidly~\cite{wanBriefSurveyRecent2023,harkatCyberphysicalSystemsSecurity2024}. 
This challenge is particularly pronounced in \textit{Operational Technology} (OT) 
environments due to their increasing connection to IT networks~\cite{maSurveyCyberSecurity2024}, 
where cyber vulnerabilities can directly impact physical processes and 
safety-critical systems~\cite{bhamareCybersecurityIndustrialControl2020}. 
Unlike traditional IT infrastructures, OT networks often rely on legacy 
protocols, require strict operational uptime, and offer limited tolerance 
for intrusive interventions. These characteristics create unique constraints 
for security analysis and risk mitigation, demanding approaches that balance 
thorough assessment with operational safety~\cite{halawighosonReviewStaticDynamic2025}.

Robotic systems represent a particular and increasingly
important class of OT environments~\cite{surve_sok_2025,yuSurveyCyberPhysical2023}. Modern robots are highly
networked cyber-physical systems deployed in domains such as industrial
automation, logistics, and autonomous service applications \cite{reaCyberPhysicalSystems2024, ferreiraSystematicLiteratureReview2023}. Middleware frameworks such as ROS and ROS2 have become a de facto standard for building distributed robotic architectures~\cite{portugalInquiringRobotOperating2025}. While this modular communication paradigm accelerates development, it also
introduces significant attack surfaces through exposed topics, services,
and nodes, as well as historically permissive default configurations \cite{bottaCyberSecurityRobots2023}.

Recent advancements in \textit{Large Language Models} (LLMs) have opened new possibilities for partially automating security tasks such as reconnaissance, vulnerability triage, exploit generation, and reporting. Their ability to generalize across tasks and adapt to diverse inputs makes them attractive for complex security workflows~\cite{goyalHackingLazyWay2025,xuAutoAttackerLargeLanguage2024,dengPentestGPTLLMempoweredAutomatic2024,isozakiAutomatedPenetrationTesting2025}. However, relying on a single LLM for penetration testing is often insufficient due to limited long-term memory, weak role separation, and reduced traceability of decisions. Furthermore, regulatory frameworks such as the EU AI Act~\cite{eu_ai_act} introduce requirements for traceability, technical documentation, and effective human oversight, particularly for safety-relevant AI systems. Purely monolithic LLM-based solutions may struggle to satisfy these requirements in practical deployments.

In this context, multi-agentic LLM-systems promise to automate complex processes such as penetration testing by orchestrating specialized agents and decomposing large challenges into manageable sub-problems. In this work, we explore how the advantages of LLMs and multi-agent systems can be leveraged while improving regulatory alignment. To this end, we propose an environment-grounded multi-agent architecture tailored for ROS-based penetration testing. Rather than relying on predefined knowledge structures, the system dynamically constructs a shared graph-based memory during execution that is anchored in observable system state. This memory incrementally captures the discovered network topology and operational artifacts, including nodes, communication channels, vulnerabilities, and attempted exploits. This approach improves coordination, reduces redundant exploration, and provides a transparent record of agent reasoning and actions. We evaluate the workflow based on a recent taxonomy~\cite{cemriWhyMultiAgentLLM2025a} for multi-agent LLM-systems and in a robotics capture-the-flag (CTF) environment. Results are compared against a literature state-of-the-art, and we additionally analyze performance across different LLM-architectures and model sizes to investigate how smaller state-of-the-art models affect effectiveness and efficiency.

% ==================================================
\section{Related Work}

Recent advances in LLM-driven offensive security have led to a new generation of semi-autonomous penetration-testing assistants. Early systems~\cite{GitHubAliasroboticsCai,happeGettingPwndAI2023} position LLMs primarily as interactive copilots that support human operators through script generation, command suggestions, and tool orchestration. These approaches focus on augmenting productivity rather than replacing human decision-making, providing recommendations that still require continuous supervision. While they demonstrate the usefulness of natural-language interfaces for security workflows, they remain fundamentally human-driven, achieving only a low degree of automation and falling short of true semi-autonomous operation. 

More recent research moves toward modular automation. PentestGPT~\cite{dengPentestGPTLLMempoweredAutomatic2024} introduces a structured pipeline in which discrete sub-tasks, such as reconnaissance, vulnerability scanning, and output interpretation, are partially automated through specialized prompts and tool integrations. This modularization represents an important step toward scalable workflow orchestration. However, the strong separation between modules makes system behavior difficult to interpret from an external perspective and limits the ability to generate new tasks dynamically based on previously discovered knowledge. In addition, reliance on external proprietary models raises concerns regarding privacy, reproducibility, and deployment in sensitive environments.

Multi-agent architectures such as HackSynth~\cite{muzsaiHackSynthLLMAgent2024} explore planner-summarizer paradigms to coordinate complex attack sequences. By delegating planning and summarization to distinct agents, these systems demonstrate emerging autonomous capabilities and improved task sequencing. Nevertheless, their reliance on summarization as the primary mechanism for context transfer can introduce information loss, which limits long-term consistency and reduces the persistence of operational knowledge across workflow steps.

Other work focuses on specialized phases of the attack lifecycle. AutoAttacker~\cite{xuAutoAttackerLargeLanguage2024}, for instance, emphasizes post-exploitation automation and demonstrates how LLMs can execute hands-on-keyboard activities without direct human input. Although this represents a significant advance toward operational autonomy within a constrained domain, the system lacks persistent memory structures that would allow multiple specialized agents to robustly share and reuse discovered information about the target system.

Across these approaches, traceability and human oversight remain only partially addressed, which poses challenges for two main reasons. First, existing implementations typically rely on logs or ad-hoc summaries that do not allow analysts to reliably reconstruct why specific actions were taken, which contextual information informed them, or how intermediate findings influenced subsequent steps. This limitation becomes particularly critical in regulated or safety-critical environments, where transparent and auditable decision-making processes are essential. Second, emerging regulatory frameworks such as the EU AI Act~\cite{eu_ai_act} introduce explicit requirements for traceability, technical documentation, and effective human oversight, which current systems only partially support.

Three major gaps emerge from the current state of the art in LLM-driven automation of penetration testing:

\begin{enumerate}
    \item \textbf{Incomplete Automation.}
    Existing solutions demonstrate promising capabilities for isolated tasks or specific phases, but they rarely generate autonomous tasks based on previously discovered findings. As a result, true end-to-end autonomous penetration-testing workflows remain a challenge.

    \item \textbf{Lack of Persistent Context Management.}
    Current approaches lack a context-management mechanism that is decoupled from the repetitive execution steps of the workflow. Architectures either risk losing information between modules or rely heavily on summarization, which introduces inconsistencies and reduces reliability during long-running engagements.

    \item \textbf{Limited Traceability and Human Oversight.}
    Robust mechanisms for traceability remain largely absent, highlighting the need for architectures that combine autonomous operation with transparent and auditable action histories to support effective human oversight.
\end{enumerate}

% ==================================================
\section{Methodology and Experimental Setup} 

\begin{figure*}
    \centering
    \includegraphics[width=0.8\linewidth]{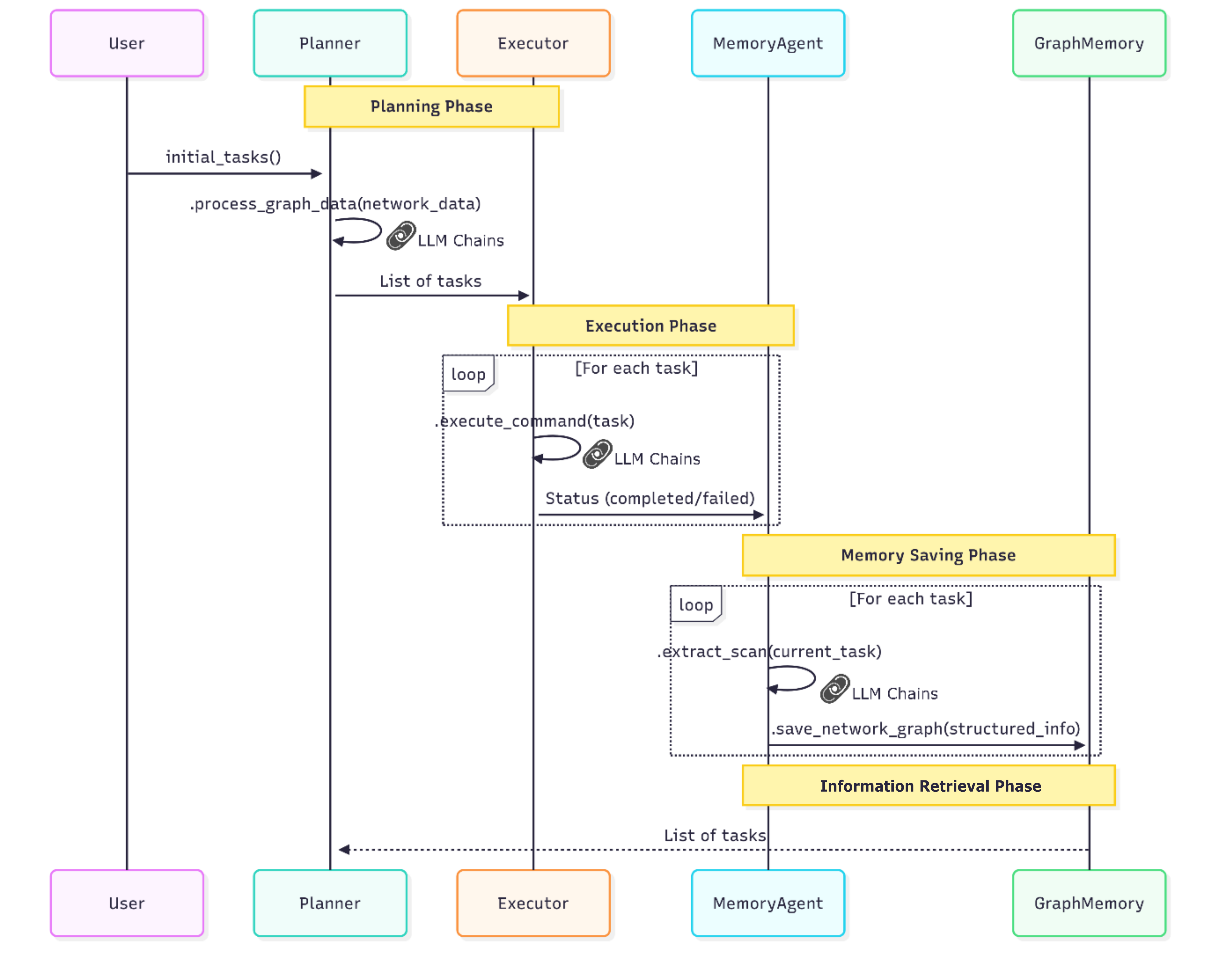}
    \caption{Overview of the agentic penetration testing workflow. The figure illustrates the interaction of the agents across the main phases of the process: task planning, task execution, memory saving, and information retrieval.}
    \label{fig:workflow}
\end{figure*}

\subsection{Proposed Agentic Penetration Testing Workflow}
The proposed solution consists of a LangGraph-based multi-agent workflow (see Fig.~\ref{fig:workflow}) with three cooperative agents operating in a closed-loop architecture: the \textit{Planner}, the \textit{Executor}, and the \textit{Memory Agent}. The penetration testing workflow is initiated by the user, typically by specifying a target range for an Nmap scan. During the engagement, the user can switch between different \textit{Operational Modes} (e.g., scanning or exploitation) to guide the assessment process.

During the \textit{Planning Phase}, the Planner generates a structured list of tasks in a predefined format. The list is validated by the workflow against the predefined syntax before being passed to the Executor. In the \textit{Execution Phase}, the Executor iterates through the tasks and translates each one into an executable command depending on the task type (e.g., Nmap scanning or ROS exploitation).

Next, in the \textit{Memory Saving Phase}, the Memory Agent reviews completed tasks and records their outcomes in the persistent knowledge base, referred to as \textit{GraphMemory}. The obtained knowledge graph represents the discovered network topology: when a new host or service is identified, it is added as a node in the graph; if the host already exists, new findings are appended using a unique task identifier. This process gradually builds a comprehensive record of executed actions, affected nodes, and resulting outcomes, enabling reconstruction of the complete attack path.

Finally, the Memory Agent queries the knowledge graph and provides relevant insights back to the Planner. By incorporating accumulated knowledge into subsequent planning cycles, the system enables iterative and context-aware progression throughout the penetration testing workflow.

The individual system components, as illustrated in Fig.~\ref{fig:workflow}, along with the various modes and concepts comprising the workflow, are described in detail below

\textbf{Operational Modes.}
In \textit{Scanning Mode}, the system performs network enumeration using Nmap, collecting information such as open ports, detected services, version information, service banners, and operating system fingerprints. When port 11311 is discovered, indicating a potential Robot Operating System (ROS) environment, the workflow switches to \textit{ROS Exploitation Mode}. In this mode, the system focuses on identifying ROS1 or ROS2 topics and extracting data from active communication channels.

\textbf{Tools.}
The Executor relies on a small set of tools to perform tasks: Nmap (CLI) for host discovery and service enumeration, and custom Bash scripts for ROS-specific exploitation activities.

\textbf{State and Task Management.}
The workflow is organized around two core concepts: \textit{State} and \textit{Task}. In LangGraph, the State acts as the shared memory passed between agents and represents the single source of truth for the current workflow execution. It tracks the current stage of execution, pending and completed tasks, and intermediate data required for reporting or visualization. This ensures that all agents remain synchronized throughout the workflow.

A Task represents a single unit of work within the workflow. Tasks are generated by the Planner, executed by the Executor, and summarized by the Memory Agent. This lifecycle ensures that each action performed during the assessment is documented and that its outcome is clearly recorded.

\textbf{Persistent Memory.}
In addition to the temporary workflow State, the system maintains a persistent knowledge base referred to as \textit{GraphMemory}. This memory captures the evolving representation of the target network throughout the engagement. For each discovered host, the system records a simplified history of executed actions, such as scans or exploitation attempts, together with their outcomes. The information is stored in a structured graph representation that continuously grows as the assessment progresses, enabling the system to maintain an up-to-date view of the discovered network topology.

With respect to automation, we intentionally adopt a middle-ground design between fully autonomous and fully manual penetration testing systems. The workflow operates within predefined operational modes (e.g., reconnaissance or exploitation), which define the scope of permissible actions and available tools. A human-in-the-loop is responsible for switching between these predefined modes. Within each mode, the agents operate autonomously: they discover information, update the shared memory, generate follow-up tasks, and iteratively refine the plan based on newly observed system states. This approach allows the system to retain the adaptive reasoning capabilities of autonomous agents while maintaining operational control and safety through mode boundaries and guardrail constraints.
The code used in this study is not publicly available for ethical reasons but is available from the corresponding author upon request.
% ==================================================

\subsection{System under Test}
\begin{figure}[h!]
    \centering
    \includegraphics[width=0.8\linewidth]{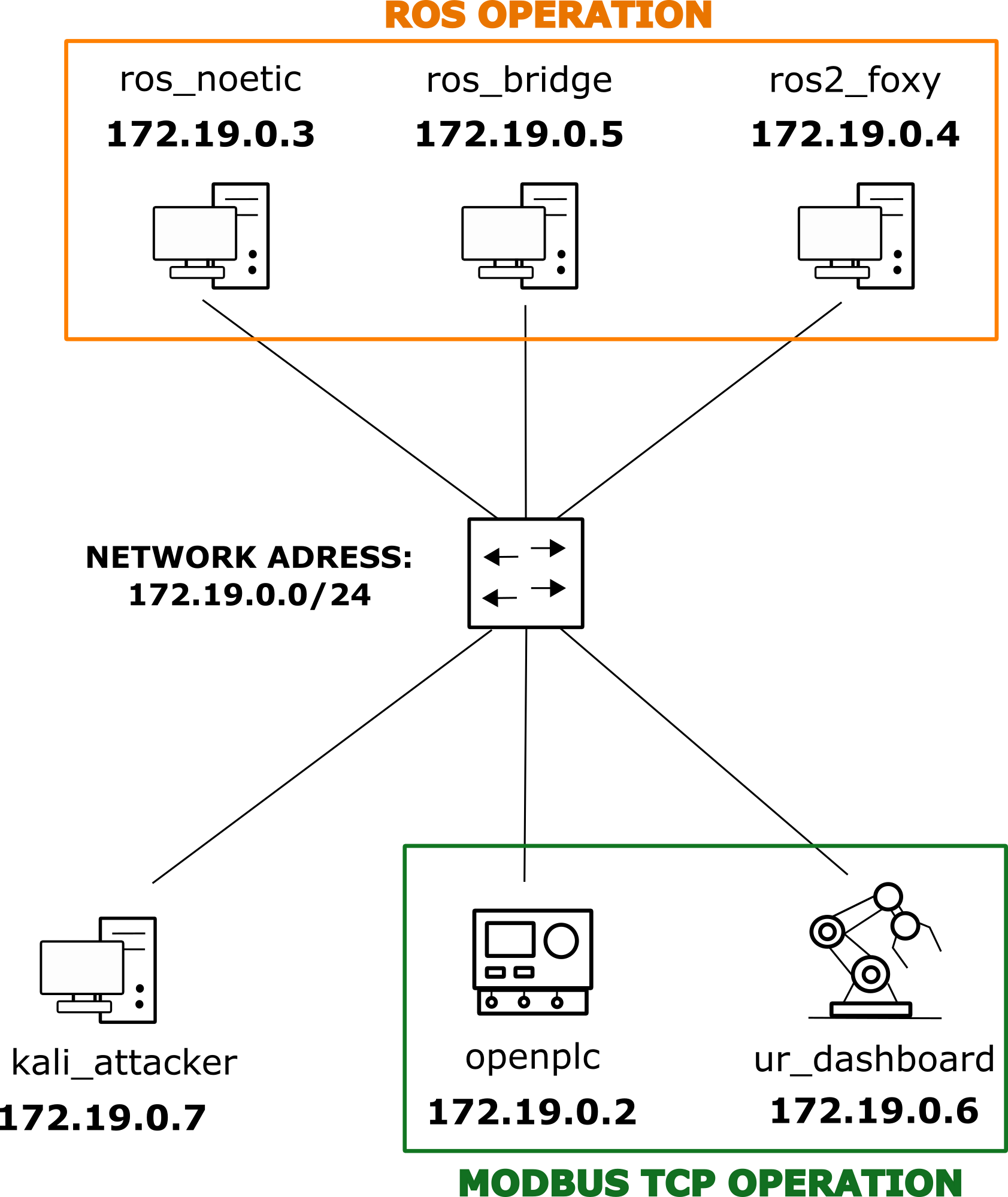}
    \caption{Docker network representing the robot-based manufacturing use case under test.}
    \label{fig:ros_docker_network}
\end{figure}

The proposed approach to automate penetration testing using LLMs were evaluated in a controlled experimental environment derived from a robot-based manufacturing use case. A representative testbed was designed to preserve the essential architectural and operational characteristics of such systems while limiting complexity to enable controlled experimentation and repeatability.

The testbed consists of the core components shown in Fig.~\ref{fig:ros_docker_network} and is implemented as a Docker-based network architecture, comprising the following elements:

\begin{itemize}
    \item \textbf{ROS-based robotic communication layer:} 
    Two ROS containers (\texttt{ros\_noetic}, \texttt{ros2\_foxy}) emulate manufacturing robots. These containers are interconnected through a third ROS container (\texttt{ros\_bridge}) that acts as a bridge, enabling communication between the two nodes via a ROS bridge mechanism. The system generates a continuous data flow based on a publisher-subscriber communication model, thereby replicating typical message exchange patterns in distributed robotic systems.
    
    \item \textbf{Programmable Logic Controller (PLC):} 
    A PLC (\texttt{openplc}) continuously transmits a data stream representing the state of a specific coil. This signal serves as a control variable within the system and mimics real-world industrial control signaling behavior.
    
    \item \textbf{Industrial robotic arm:} 
    A Universal Robots (UR) robotic arm (\texttt{ur\_dashboard}) performs cyclic motion, simulating realistic manufacturing activity. The robot’s operation depends on the continuous data stream received from the PLC.
\end{itemize}

The containerized environment closely emulates the real-world networking and message-passing mechanisms present in industrial robotics and automation systems. In this study, the security analysis is primarily focused on the ROS-based communication layer (see the ROS OPERATION box in Fig.~\ref{fig:ros_docker_network}), as it constitutes a critical and externally exposed component of the overall architecture.

ROS-based real world deployments typically rely on unencrypted and unauthenticated TCP/IP communication by default, thereby introducing significant security risks~\cite{bottaCyberSecurityRobots2023}. 
In the case of \texttt{ros\_noetic} (ROS~1), the absence of built-in authentication and encryption mechanisms results in unauthenticated open ports, plaintext message exchange, susceptibility to node spoofing, and unrestricted topic enumeration. Although \texttt{ros2\_foxy} (ROS~2) adopts a DDS-based communication model, security mechanisms such as encryption and authentication remain optional and are not always enabled in practice. When misconfigured, this may allow message sniffing, denial-of-service (DoS) attacks, and message injection.

%---------------------------------------------------
\subsection{Evaluation} 

We employ a Capture-the-Flag (CTF)-style evaluation to assess penetration testing performance across different attack phases. A CTF challenge is a cybersecurity exercise in which participants solve security-related tasks—such as exploiting vulnerabilities, reverse engineering, or cryptography—to retrieve hidden 'flags.' This format serves as a suitable benchmark because it provides a realistic, hands-on simulation of adversarial scenarios and enables the evaluation of practical offensive security skills in a controlled and repeatable environment~\cite{mitll2016ctf}.

\textbf{Reconnaissance Phase.}  
CTF-0 requires the discovery of all six nodes within the Docker network.  
CTF-1 focuses on identifying the ROS master:
\texttt{RM\_WP6\_CTF\{open\_port\_11311\_on\_172.19.0.3\}}.  
Successfully obtaining this flag reflects the completion of the reconnaissance phase, as detecting the ROS master port constitutes the primary entry point into the ROS~1 network.

\textbf{Exploitation Phase.}  
CTF-2 evaluates the ability to enumerate active ROS topics:
\texttt{RM\_WP6\_CTF\{topics:}\texttt{chatter,}\texttt{rosout,} \texttt{rosout\_agg\}}. CTF-3 requires reading an active topic message:
\texttt{RM\_WP6\_CTF\{Hello\_from\_ROS\_1\}}.  
This final objective represents successful exploitation and data access, demonstrating the transition from network discovery to extracting live runtime information.

The evaluation of the proposed multi-agentic workflow is grounded in the recently proposed Multi-Agent System Failure Taxonomy (MAST), introduced in~\cite{cemriWhyMultiAgentLLM2025a}. MAST was developed through a systematic analysis of 150 execution traces, guided by expert human annotators and validated through high inter-annotator agreement. This taxonomy provides a structured framework for categorizing reasoning, planning, and coordination failures in LLM-based multi-agent systems.

To ensure a diverse and representative evaluation across different providers and model scales, we experiment with multiple planner models: \texttt{llama-3.3-70b-instruct}, \texttt{deepseek-v3.2}, \texttt{gemma-3-27b-it}, and \texttt{hermes-2-pro-llama-3-8b}. 

HackSynth~\cite{muzsaiHackSynthLLMAgent2024} adopts a planner–summarizer paradigm to coordinate complex attack sequences by delegating reasoning and context management to separate agents within a multi-agent system. This design enables a degree of autonomous behavior and structured task execution, making HackSynth a suitable benchmark for our study. To ensure a fair comparison, we follow the HackSynth setup by varying only the \textit{Planner} component across experiments, while keeping the Executor and Memory Saver fixed (implemented using \texttt{llama-3.1-8b-instruct}). Each planner model is evaluated over five independent runs to account for stochastic variability.

% ==================================================
\section{Results and Discussion}
\subsection{The Proposed Multi-Agent Workflow in Operation}

The workflow is typically initiated in \textit{Scanning Mode}, where the user provides the Planner with a structured task list for host discovery and service enumeration. An example of such a task is shown in Listing~\ref{lst:initial_task}.

\begin{lstlisting}[style=json,caption={Initial scanning task generated by the Planner.},label={lst:initial_task}]
{
  "task_id": 1,
  "description": "Perform a basic network discovery scan to identify active hosts, services and ports",
  "target": "172.19.0.0/24",
  "dependencies": [],
  "task_type": "nmap_scan"
}
\end{lstlisting}

The Executor processes the first task from the Planner's list, marking the beginning of active interaction with the target environment. It translates the task into a concrete command and executes it ([Execution Phase]). Once the command finishes, the MemoryAgent structures the output and stores the extracted information in the GraphMemory ([Memory Saving Phase]).

At this early stage, when no nodes may yet exist, the MemoryAgent creates new graph entries and records which task discovered which information on which node. This ensures that the system maintains a consistent and evolving representation of the network state. An example of a GraphMemory entry is shown in Listing~\ref{lst:graph_entry}.

\begin{lstlisting}[style=json,caption={Example GraphMemory entry created after host discovery.},label={lst:graph_entry}]
{
  "172.19.0.1": [
    {
      "id": 1,
      "task_type": "nmap_scan",
      "instruction": "Perform a basic network discovery scan",
      "memory": "Node discovered via Nmap range scan"
    }
  ]
}
\end{lstlisting}

Next, the MemoryAgent retrieves the stored information and provides it to the Planner in what can be considered the [Information Retrieval Phase]. Based on this context, the Planner enters the \textit{Planning Phase} and generates a new set of tasks for the next iteration. A snippet of an example tasks is shown in Listing~\ref{lst:planner_tasks}.

\begin{lstlisting}[style=json,caption={Example follow-up tasks generated by the Planner.},label={lst:planner_tasks}]
[
  {
    "task_id": 2,
    "description": "Conduct an intensive scan on node 172.19.0.3 to identify open ports and services",
    "target": "172.19.0.3",
    "dependencies": [1],
    "task_type": "nmap_scan"
  }
]
\end{lstlisting}

The results of these tasks are again processed by the MemoryAgent and integrated into the GraphMemory. Since the node already exists in the graph, the new task entry together with its summarized results is appended to the stored network view.

When the user switches to a different Mode, like the \textit{ROS Exploitation Mode}, the MemoryAgent loads the network information collected during the scanning phase and provides it to the Planner. Based on this context, the Planner generates exploitation tasks targeting vulnerable services. An example task is shown in Listing~\ref{lst:ros_task}.

\begin{lstlisting}[style=json,caption={Example ROS exploitation task generated by the Planner.},label={lst:ros_task}]
[
  {
    "task_id": 1,
    "description": "Exploit ROS1 vulnerability on host with open port 11311",
    "target": "172.19.0.3",
    "dependencies": [],
    "task_type": "ros_exploit"
  }
]
\end{lstlisting}

Finally, the Executor converts these tasks into executable Bash commands and runs them against the target system. Then again the MemoryAgent structures the information into the \textit{GraphMemory}. After that the infromation is passed back to the Planner and the next Plannign step beginns.

% --------------------------------------------------------------------------------%
\subsection{Literature Comparison and Performance}
\begin{table*}[t]
\centering
\footnotesize
\caption{Performance of the evaluated LLMs across CTF tasks over five attempts per task, reported as (success, failure) counts.}
\label{tab:ctf_results}
\begin{tabular}{l l | c c c c}
\toprule
Approach & Model & CTF-0 & CTF-1 & CTF-2 & CTF-3 \\
\midrule
\multirow{4}{*}{Proposed}
 & \texttt{llama-3.3-70b-instruct} & (5, 0) & (5, 0) & (5, 0) & (5, 0) \\
 & \texttt{deepseek-v3.2} & (5, 0) & (1, 4) & (1, 4) & (0, 5) \\
 & \texttt{gemma-3-27b-it} & (5, 0)  & (0, 5)  & (0, 5) & (0, 5) \\
 & \texttt{hermes-2-pro-llama-3-8b} & (5, 0)  & (0, 5) & (0, 5) & (0, 5) \\
\midrule
\multirow{4}{*}{HackSynth~\cite{muzsaiHackSynthLLMAgent2024}}
 & \texttt{llama-3.3-70b-instruct} & (5, 0) & (4, 1) & (0, 5) & (0, 5) \\
 & \texttt{deepseek-v3.2} & (4, 1) & (2, 3)  & (0, 5) & (0, 5) \\
 & \texttt{gemma-3-27b-it} & (4, 1) & (1, 4) &  (0, 5) &  (0, 5)\\
 & \texttt{hermes-2-pro-llama-3-8b} & (4, 1) & (0, 5) & (0, 5) & (0, 5) \\
\bottomrule
\end{tabular}
\end{table*}

Table~\ref{tab:ctf_results} summarizes the performance of the evaluated LLMs on the CTF tasks across five attempts per task. The outcome of each attempt is represented as tuples (success, failure) in the table. We begin by reporting the evaluation results obtained using the HackSynth~\cite{muzsaiHackSynthLLMAgent2024} setup. The results indicate that the basic network discovery task (CTF-0), which requires identifying nodes within the network, can be solved reliably. Stronger models generally complete this task consistently, often using similar command sequences across attempts. Among the evaluated models, \texttt{llama-3.3-70b-instruct} demonstrates the most stable performance. However, the results change significantly for the more specialized tasks. While CTF-1 can still be partially solved, the ROS-specific challenges (CTF-2 and CTF-3) remain unsolved across all evaluated models. This occurs despite the prompts providing the same contextual information indicating that the environment is a robotics system. During the experiments, several recurring limitations were observed. First, task descriptions must closely match the expected solution space; overly open-ended instructions often lead to incorrect or irrelevant command generation. For example, when generating \texttt{nmap} commands, the model often produces an excessive number of flags (e.g., timeouts, scan optimizations, or aggressive options), resulting in overly complex or suboptimal commands that deviate from the intended task. Second, models often generate actions that do not fully correspond to their reasoning process. For example, \texttt{nmap} scans are frequently executed with default parameters, probing only commonly used ports rather than performing a full port scan. As a result, relevant services remain undiscovered and the attack path is lost. This behaviour aligns with the \textit{Reasoning-Action Mismatch} failure category (2.6)~\cite{cemriWhyMultiAgentLLM2025a}. Third, models sometimes repeat similar scanning attempts or continue exploring ineffective strategies instead of adapting their approach after failed attempts. This behaviour corresponds to \textit{Step Repetition} (1.3), where agents repeatedly execute similar actions without progressing towards task completion~\cite{cemriWhyMultiAgentLLM2025a}.

Overall, the experiments with the HackSynth~\cite{muzsaiHackSynthLLMAgent2024} framework highlight three key challenges for state-of-the-art autonomous LLM-based penetration testing systems: limited controllability of model behavior, insufficient traceability of generated actions for users and compliance requirements (e.g., documentation obligations under the AI Act), and the difficulty of handling domain-specific security tasks, such as ROS penetration testing, using general-purpose LLM-based agents.

%%%%%%%%%%%%%%%%%%%

Table~\ref{tab:ctf_results} also reports the performance of the proposed multi-agent workflow. In contrast to the literature baseline, the proposed system successfully completes all reconnaissance tasks and most exploitation tasks. In the reconnaissance phase (CTF-0 and CTF-1), the workflow reliably identifies all active nodes and the ROS master port, which serves as the entry point into the robotic network.

For CTF-0, all evaluated models successfully discover the six active nodes in the network. In CTF-1, \texttt{llama-3.3-70b-instruct} consistently identifies the ROS master port (11311) across all attempts. While \texttt{deepseek-v3.2} correctly discovers all active nodes, it fails to detect the ROS-specific service despite the relevant port being present in the scan results. Similarly, \texttt{gemma-3-27b-it} performs extensive scanning but produces overly verbose reasoning and diverges from the intended task by initiating additional vulnerability scans. This behaviour aligns with \textit{Task Derailment} (2.3), where agents deviate from the assigned objective and pursue unrelated actions~\cite{cemriWhyMultiAgentLLM2025a}. The \texttt{hermes-2-pro-llama-3-8b} model executes syntactically valid commands but prematurely restricts scanning to specific TCP ports, resulting in no discovered services. This behaviour can be interpreted as a \textit{Reasoning--Action Mismatch} (2.6), where the executed actions do not effectively implement the intended exploration strategy.

In the exploitation phase (CTF-2 and CTF-3), the workflow demonstrates the advantage of combining LLM-based planning with deterministic rule-based components. A predefined routine captures relevant console output generated by the Executor and detects ROS topic enumeration results. This mechanism enables seamless integration between unstructured LLM reasoning and structured script-based validation.

As a result, both \texttt{llama-3.3-70b-instruct} and \texttt{deepseek-v3.2} successfully enumerate ROS topics and retrieve the target message, solving CTF-2 and CTF-3. Models that failed during the reconnaissance phase (i.e., \texttt{gemma-3-27b-it} and \texttt{hermes-2-pro-llama-3-8b}) were unable to proceed to the exploitation stage due to the missing discovery of the ROS master port.

Beyond performance improvements, the proposed workflow also improves system-level properties. First, controllability can be enforced through structured constraints on agent actions, such as restricting valid IP ranges and tool usage through predefined configuration matrices. Second, the GraphMemory enables full traceability of the attack workflow by recording which agent executed which command at which stage of the process. This allows reconstruction of the attack graph and improves transparency for users, which is particularly relevant for regulatory requirements such as documentation obligations under the AI Act in high-stakes security applications.

Nevertheless, some limitations remain. Agents occasionally generate overly complex commands, which increases scan duration and reduces efficiency. This issue could be mitigated by incorporating tool documentation and usage guidelines directly into the planning phase. 

%%%%%%%%%%%%%%%%%%%%%%%%%
% ==================================================
\section{Conclusion}

This paper proposes a multi-agent workflow utilizing LangGraph, a state-of-the-art orchestration framework. By deploying the system on a ROS-based OT network, we demonstrated that our approach, which leverages environment-grounded memory, can significantly outperform existing benchmarks reported in the literature. Beyond raw performance, the system was designed with controllability, human oversight, and comprehensive traceability as core principles. These design considerations address both the practical demands of industrial deployment and the stringent requirements of emerging regulatory frameworks for autonomous systems.
Looking ahead, we aim to further advance persistent memory mechanisms in multi-agent workflows. We believe this paradigm has significant potential for application in other safety-critical domains, such as critical infrastructure monitoring and industrial process control.
% ==================================================
\section*{Acknowledgments}

This work was funded by the European Union's Horizon Research and Innovation Programme under GA No. 101119681 (project ResilMesh). Views and opinions expressed are however those of the author(s) only and do not necessarily reflect those of the European Union or the European Commission. Neither the European Union nor the granting authority can be held responsible for them.

% Generated by IEEEtran.bst, version: 1.14 (2015/08/26)


\begin{thebibliography}{10}
\providecommand{\url}[1]{#1}
\csname url@samestyle\endcsname
\providecommand{\newblock}{\relax}
\providecommand{\bibinfo}[2]{#2}
\providecommand{\BIBentrySTDinterwordspacing}{\spaceskip=0pt\relax}
\providecommand{\BIBentryALTinterwordstretchfactor}{4}
\providecommand{\BIBentryALTinterwordspacing}{\spaceskip=\fontdimen2\font plus
\BIBentryALTinterwordstretchfactor\fontdimen3\font minus \fontdimen4\font\relax}
\providecommand{\BIBforeignlanguage}[2]{{%
\expandafter\ifx\csname l@#1\endcsname\relax
\typeout{** WARNING: IEEEtran.bst: No hyphenation pattern has been}%
\typeout{** loaded for the language `#1'. Using the pattern for}%
\typeout{** the default language instead.}%
\else
\language=\csname l@#1\endcsname
\fi
#2}}
\providecommand{\BIBdecl}{\relax}
\BIBdecl

\bibitem{wanBriefSurveyRecent2023}
Y.~Wan and J.~Cao, ``\BIBforeignlanguage{en}{A {Brief} {Survey} of {Recent} {Advances} and {Methodologies} for the {Security} {Control} of {Complex} {Cyber}–{Physical} {Networks}},'' \emph{\BIBforeignlanguage{en}{Sensors}}, vol.~23, no.~8, Apr. 2023.

\bibitem{harkatCyberphysicalSystemsSecurity2024}
H.~Harkat, L.~M. Camarinha-Matos, J.~Goes, and H.~F.~T. Ahmed, ``Cyber-physical systems security: {A} systematic review,'' \emph{Computers \& Industrial Engineering}, vol. 188, p. 109891, Feb. 2024.

\bibitem{maSurveyCyberSecurity2024}
Y.-W. Ma, Y.-H. Tu, C.-W. Tsou, Y.-N. Chiang, and J.-L. Chen, ``\BIBforeignlanguage{en-US}{A {Survey} of {Cyber} {Security} and {Safety} in {Industrial} {Control} {Systems}},'' \emph{\BIBforeignlanguage{en-US}{Journal of Internet Technology}}, vol.~25, no.~4, pp. 541--550, Jul. 2024.

\bibitem{bhamareCybersecurityIndustrialControl2020}
D.~Bhamare, M.~Zolanvari, A.~Erbad, R.~Jain, K.~Khan, and N.~Meskin, ``Cybersecurity for industrial control systems: {A} survey,'' \emph{Computers \& Security}, vol.~89, p. 101677, Feb. 2020.

\bibitem{halawighosonReviewStaticDynamic2025}
N.~Halawi~Ghoson, V.~Meyrueis, K.~Benfriha, T.~Guiltat, and S.~Loubère, ``A review on the static and dynamic risk assessment methods for {OT} cybersecurity in industry 4.0,'' \emph{Computers \& Security}, vol. 150, p. 104295, Mar. 2025.

\bibitem{surve_sok_2025}
\BIBentryALTinterwordspacing
P.~P. Surve, A.~Shabtai, and Y.~Elovici, ``{SoK}: {Cybersecurity} {Assessment} of {Humanoid} {Ecosystem},'' Sep. 2025. [Online]. Available: \url{http://arxiv.org/abs/2508.17481}
\BIBentrySTDinterwordspacing

\bibitem{yuSurveyCyberPhysical2023}
Z.~Yu, H.~Gao, X.~Cong, N.~Wu, and H.~H. Song, ``A {Survey} on {Cyber}–{Physical} {Systems} {Security},'' \emph{IEEE Internet of Things Journal}, vol.~10, no.~24, pp. 21\,670--21\,686, Dec. 2023.

\bibitem{reaCyberPhysicalSystems2024}
P.~Rea, M.~Ruggiu, P.~Ponchietti, E.~Ottaviano, and A.~G. Gonzalez~Rodriguez, ``Cyber {Physical} {Systems}: {A} {Brief} {Survey} and an {Application} of a {MIR} ({Mobile} {Industrial} {Robot}) for {Inspection},'' \emph{IFAC-PapersOnLine}, vol.~58, no.~8, pp. 246--251, Jan. 2024.

\bibitem{ferreiraSystematicLiteratureReview2023}
B.~Ferreira and J.~Reis, ``\BIBforeignlanguage{en}{A {Systematic} {Literature} {Review} on the {Application} of {Automation} in {Logistics}},'' \emph{\BIBforeignlanguage{en}{Logistics}}, vol.~7, no.~4, Nov. 2023.

\bibitem{portugalInquiringRobotOperating2025}
D.~Portugal, R.~P. Rocha, and J.~P. Castilho, ``\BIBforeignlanguage{en}{Inquiring the robot operating system community on the state of adoption of the {ROS} 2 robotics middleware},'' \emph{\BIBforeignlanguage{en}{International Journal of Intelligent Robotics and Applications}}, vol.~9, no.~2, pp. 454--479, Jun. 2025.

\bibitem{bottaCyberSecurityRobots2023}
A.~Botta, S.~Rotbei, S.~Zinno, and G.~Ventre, ``Cyber security of robots: {A} comprehensive survey,'' \emph{Intelligent Systems with Applications}, vol.~18, p. 200237, May 2023.

\bibitem{goyalHackingLazyWay2025}
D.~Goyal, S.~Subramanian, A.~Peela, and N.~P. Shetty, ``Hacking, {The} {Lazy} {Way}: {LLM} {Augmented} {Pentesting},'' May 2025.

\bibitem{xuAutoAttackerLargeLanguage2024}
J.~Xu, J.~W. Stokes, G.~McDonald, X.~Bai, D.~Marshall, S.~Wang, A.~Swaminathan, and Z.~Li, ``{AutoAttacker}: {A} {Large} {Language} {Model} {Guided} {System} to {Implement} {Automatic} {Cyber}-attacks,'' Mar. 2024.

\bibitem{dengPentestGPTLLMempoweredAutomatic2024}
G.~Deng, Y.~Liu, V.~Mayoral-Vilches, P.~Liu, Y.~Li, Y.~Xu, T.~Zhang, Y.~Liu, M.~Pinzger, and S.~Rass, ``{PentestGPT}: {An} {LLM}-empowered {Automatic} {Penetration} {Testing} {Tool},'' Jun. 2024.

\bibitem{isozakiAutomatedPenetrationTesting2025}
I.~Isozaki, M.~Shrestha, R.~Console, and E.~Kim, ``Towards {Automated} {Penetration} {Testing}: {Introducing} {LLM} {Benchmark}, {Analysis}, and {Improvements},'' Feb. 2025.

\bibitem{eu_ai_act}
\BIBentryALTinterwordspacing
{European Parliament and Council of the European Union}, \emph{Regulation ({EU}) 2024/1689 of the {European} {Parliament} and of the {Council} laying down harmonised rules on artificial intelligence ({Artificial} {Intelligence} {Act})}, ser. Official {Journal} of the {European} {Union}, 2024. [Online]. Available: \url{https://eur-lex.europa.eu/}
\BIBentrySTDinterwordspacing

\bibitem{cemriWhyMultiAgentLLM2025a}
M.~Cemri, M.~Z. Pan, S.~Yang, L.~A. Agrawal, B.~Chopra, R.~Tiwari, K.~Keutzer, A.~Parameswaran, D.~Klein, K.~Ramchandran, M.~Zaharia, J.~E. Gonzalez, and I.~Stoica, ``Why {Do} {Multi}-{Agent} {LLM} {Systems} {Fail}?'' Oct. 2025.

\bibitem{GitHubAliasroboticsCai}
\BIBentryALTinterwordspacing
``{GitHub} - aliasrobotics/cai: {Cybersecurity} {AI} ({CAI}), the framework for {AI} {Security}.'' [Online]. Available: \url{https://github.com/aliasrobotics/cai}
\BIBentrySTDinterwordspacing

\bibitem{happeGettingPwndAI2023}
A.~Happe and J.~Cito, ``\BIBforeignlanguage{en}{Getting pwn’d by {AI}: {Penetration} {Testing} with {Large} {Language} {Models}},'' in \emph{\BIBforeignlanguage{en}{Proceedings of the 31st {ACM} {Joint} {European} {Software} {Engineering} {Conference} and {Symposium} on the {Foundations} of {Software} {Engineering}}}.\hskip 1em plus 0.5em minus 0.4em\relax San Francisco CA USA: ACM, Nov. 2023, pp. 2082--2086.

\bibitem{muzsaiHackSynthLLMAgent2024}
\BIBentryALTinterwordspacing
L.~Muzsai, D.~Imolai, and A.~Lukács, ``{HackSynth}: {LLM} {Agent} and {Evaluation} {Framework} for {Autonomous} {Penetration} {Testing},'' Dec. 2024, arXiv:2412.01778 [cs]. [Online]. Available: \url{http://arxiv.org/abs/2412.01778}
\BIBentrySTDinterwordspacing

\bibitem{mitll2016ctf}
{MIT Lincoln Laboratory}, ``Can a game teach practical cyber security?'' \emph{Lincoln Laboratory Journal}, vol.~22, no.~1, pp. 16--18, 2016.

\end{thebibliography}
\end{document}